\documentclass{article}

\usepackage{microtype}
\usepackage{graphicx}
\usepackage{subfigure}
\usepackage{booktabs} 

\usepackage{hyperref}

\usepackage[accepted]{icml2024}

\usepackage{amsmath}
\usepackage{amssymb}
\usepackage{mathtools}
\usepackage{amsthm}

\usepackage[capitalize,noabbrev]{cleveref}

\theoremstyle{plain}

\theoremstyle{definition}

\theoremstyle{remark}

\newcommand{\model}{CoIn}

\usepackage{colortbl}
\newcolumntype{G}{>{\columncolor{gray!25}}c}
\colorlet{shadecolor}{gray!30}
\definecolor{graycell}{gray}{0.9} 
\usepackage{kotex}
\usepackage{bm}
\usepackage{multirow}

\usepackage{titletoc}

\icmltitlerunning{Adapting Pretrained ViTs with Convolution Injector for Visuo-Motor Control}

\begin{document}

\twocolumn[
\icmltitle{Adapting Pretrained ViTs with Convolution Injector for Visuo-Motor Control}



\icmlsetsymbol{equal}{*}

\begin{icmlauthorlist}
\icmlauthor{Dongyoon Hwang}{equal,sch}
\icmlauthor{Byungkun Lee}{equal,sch}
\icmlauthor{Hojoon Lee}{sch}
\icmlauthor{Hyunseung Kim}{sch}
\icmlauthor{Jaegul Choo}{sch}
\end{icmlauthorlist}

\icmlaffiliation{sch}{Kim Jaechul Graduate School of AI, KAIST}
\icmlcorrespondingauthor{Dongyoon Hwang}{godnpeter@kaist.ac.kr}

\icmlkeywords{Machine Learning, ICML}

\vskip 0.3in
]



\printAffiliationsAndNotice{\icmlEqualContribution} 

\begin{abstract}
Vision Transformers (ViT), when paired with large-scale pretraining, have shown remarkable performance across various computer vision tasks, primarily due to their weak inductive bias.
However, while such weak inductive bias aids in pretraining scalability, this may hinder the effective adaptation of ViTs for visuo-motor control tasks as a result of the absence of control-centric inductive biases.
Such absent inductive biases include spatial locality and translation equivariance bias which convolutions naturally offer. To this end, we introduce \textbf{Co}nvolution \textbf{In}jector  (\textbf{{\model}}), an add-on module that injects convolutions which are rich in locality and equivariance biases into a pretrained ViT for effective adaptation in visuo-motor control. We evaluate {\model} with three distinct types of pretrained ViTs (CLIP, MVP, VC-1) across 12 varied control tasks within three separate domains (Adroit, MetaWorld, DMC), and demonstrate that {\model} consistently enhances control task performance across all experimented environments and models, validating the effectiveness of providing pretrained ViTs with control-centric biases.\footnote{\footnotesize Project page: \parbox[t]{0.5\textwidth}{\url{https://godnpeter.github.io/CoIn}}}\footnote{\footnotesize Code: \url{https://github.com/dojeon-ai/CoIn}}
\end{abstract}

\section{Introduction}

Developing intelligent robotic agents capable of precise visuo-motor control is an important area of research.
A standard paradigm of developing such agents is to train the visual encoder and control policy end-to-end, using domain-specific control data~\cite{levine2016e2e_visual}.
However, this approach limits the applicability of visuo-motor control policies in real-world scenarios due to the excessive amount of data required for learning and the lack of flexibility in adapting to new situations, such as unseen environments. 

In the fields of computer vision and natural language processing, a large number of studies have shown that pretraining high-capacity models on large datasets demonstrate superior data efficiency and generalization capabilities compared to approaches trained from scratch \citep{dosovitskiy2020image, bommasani2021foundationmodels, devlin2018bert,brown2020gpt3}.
In response, for visuo-motor control, there has been a growing interest in utilizing large visual encoders pretrained on extensive and diverse datasets ~\citep{radosavovic2022mvp, hansen2021svea,majumdar2023vc1}.

\begin{figure}[t]
\begin{center}
\includegraphics[width=0.5\textwidth]{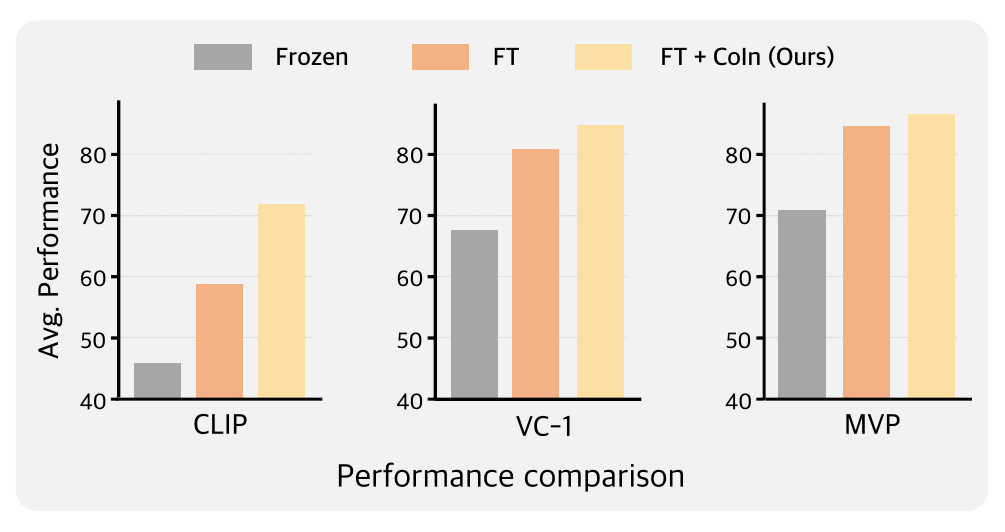}
\end{center}
\vspace{-2mm}
\caption{\textbf{Avg. performance across 12 visuo-motor control tasks.} Our model {\model} introduces convolutional inductive biases into ViTs, resulting in consistent performance improvements for various pretrained ViTs.
}
\label{fig:fig1}
\end{figure}

\begin{figure}[t]
\begin{center}
\includegraphics[width=0.46\textwidth]{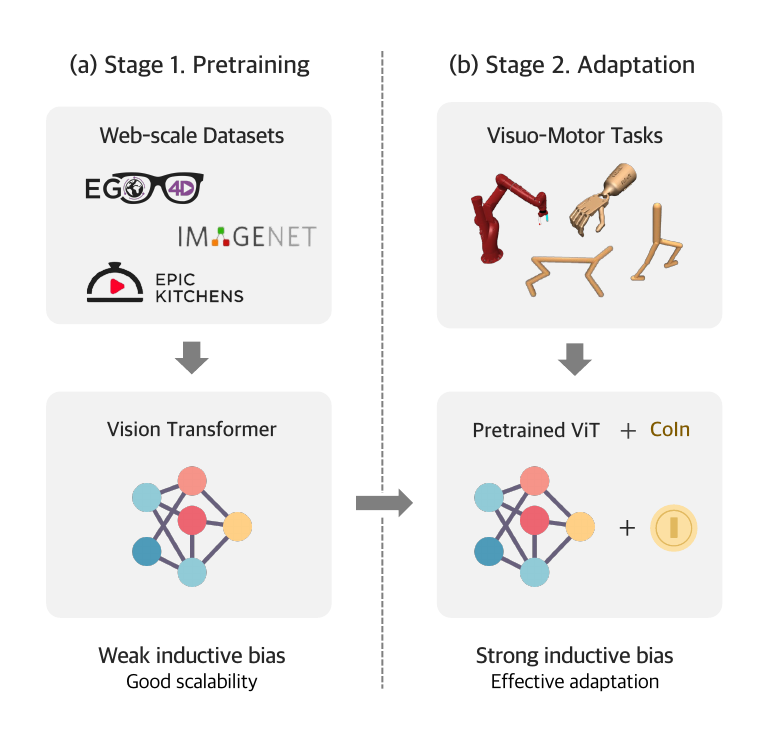}
\end{center}
\vspace{-2mm}
\caption{\textbf{Overall framework.} (Stage 1) The advent of open-sourced, large-scale ViTs pretrained with extensive web-scale datasets provides generalized, ready-to-go visual representations. (Stage 2) To adapt these pretrained ViTs for visuo-motor control, we finetune them with an additional light-weight module, {\model}, enhancing the ViT's ability to extract visual features beneficial for control, such as spatial locality and translation equivariance.}
\label{fig:main framework}
\end{figure}

For visuo-motor control, Vision Transformers (ViT)~\citep{dosovitskiy2020image} emerges as an appealing choice as it achieved remarkable success in a wide range of computer vision tasks such as image classification~\citep{dosovitskiy2020image, bao2021beit}, object detection~\citep{liu2021swin, li2022ViTDet} and semantic segmentation~\citep{strudel2021segmenter, kirillov2023sam}. 
The success of ViTs is attributed to their weak inductive bias, which significantly enhances model performance when scaled with a large pretraining dataset and model size~\citep{naseer2021intriguing,yu2021rethinking,mao2022towards,chu2021twins, dehghani2023scalingvit}.

Nonetheless, although the weak inductive bias of ViTs is advantageous for scaling during the pretraining phase, this characteristic may hinder their effective adaptation for visuo-motor control.
For effective visuo-motor control, a visual encoder must (i) focus on the interaction area of interest, and (ii) track object and gripper positions with respect to their changes in locations.
ViTs inherently lack such properties due to their design.
Such limitations of ViTs can be addressed by incorporating two specific biases that are naturally present in convolutional layers: (i) a bias towards spatial locality, and (ii) a bias for translation equivariance.

To this end,  we introduce \textbf{Co}nvolution \textbf{In}jector  (\textbf{{\model}}), a module designed to inject spatial locality and translation equivariance biases into a pretrained ViT for effective adaptation in visuo-motor control.
{\model} is a simple and lightweight add-on module (3.6\% of additional parameters to a standard ViT-B/16) designed to exploit the strengths of pretrained ViTs while providing advantageous inductive biases essential for visual control tasks.
Specifically, {\model} extracts locality and translation equivariance-aware features through convolutional layers and integrates them into the ViT architecture using a cross-attention mechanism (see Figure~\ref{fig:architecture}).
This integration enables the pretrained ViT to effectively leverage both its pretrained knowledge and newly obtained spatial prior features during adaptation for downstream control tasks.
Therefore, {\model} eliminates the need to retrain pretrained ViTs from scratch with datasets and objectives specifically tailored for visual control applications.

To thoroughly evaluate the effectiveness of {\model}, we conduct extensive evaluations across 12 different visuo-motor control tasks within 3 distinct domains: Adroit~\citep{rajeswaran2018adroit}, MetaWorld~\citep{yu2020metaworld}, and DMC~\citep{Tassa2018DMC} for three different pretrained ViT visual encoders: CLIP~\citep{radford2021clip}, MVP~\citep{radosavovic2022mvp}, and VC-1~\citep{majumdar2023vc1}. 
Our results demonstrate that {\model} consistently enhances downstream control task performance across all environments and with all pretrained ViTs. Notably, when paired with CLIP, finetuning with {\model} achieved a substantial 11.3 point increase in mean success over finetuning the baseline CLIP model.
These findings suggest that the incorporation of locality and translation-equivariance-aware features plays a crucial role in enhancing the capabilities of ViTs for visuo-motor control tasks.

In summary, although ViTs gain advantages from large-scale pretraining due to their weak inductive bias, this same characteristic limits their adaptability for visuo-motor control tasks because of the absence of specific control-centric biases. Consequently, we introduce {\model}, a module which incorporates beneficial control-centric inductive biases, readily provided by convolutional layers, into large-scale pretrained ViTs. Our code is available at \url{https://godnpeter.github.io/CoIn}.

\section{Related Work}

\subsection{Pretrained Visual Encoders for Control}
\label{subsection:2.1}

Recently, pretraining effective visual encoders for control by leveraging large, diverse datasets from the internet has gain much interest from the research community~\citep{parisi2022unsurprising, nair2022r3m, radosavovic2022mvp, majumdar2023vc1, wang2022vrl3, yuan2022pieg, shah2021rrl}. Specifically, PVR~\citep{parisi2022unsurprising} is among the initial investigations into the use of large pretrained visual encoders for control. It demonstrates that while doing behavior cloning, ResNet encoders~\citep{he2016resnet} trained via self-supervised contrastive learning~\citep{he2020moco} can match the performance of state-based inputs.
Further advancements are seen in R3M~\citep{nair2022r3m}, which employs a temporal contrastive objective to learn representations for robotic control and VIP~\citep{ma2022vip}, which focuses on learning visual representations which reflect the distance between states and goals. Similarly, MVP~\citep{radosavovic2022mvp} and VC-1~\citep{majumdar2023vc1} demonstrate the efficacy of ViTs pretrained with MAE~\citep{he2022MAE} on extensive internet video and image data for robotic manipulations. As an alternative attempt, MOO~\citep{stone2023moo} and RT-2~\citep{brohan2023rt2} investigate the application of vision-language models pretrained on broad internet data, for improved robotic control and emergent reasoning. Unlike previous research which mainly focus on the performance of \textit{frozen} weights in different control tasks, our work delves into the effectiveness and challenges of \textit{finetuning} ViTs for control tasks, particularly in the context of imitation learning.

\subsection{Integration of CNNs with Pretrained ViTs in Computer Vision}
\label{subsection:cnnplutvit}

The integration of CNNs with pretrained ViTs to leverage their collective capabilities for various computer vision tasks has recently been investigated by the research community~\cite{peng2021conformer,Fang2022UnleashingVV,chen2022vit_adapter,Ranftl2021dpt,Hong2022RepresentationSF}.
VitMatte~\cite{yao2024vitmattte} demonstrates the effectiveness of combining lightweight CNNs with a pretrained ViT for enhanced image matting. 
For dense prediction, DPT~\cite{Ranftl2021dpt} introduces a randomly initialized CNN decoder, and ViT-Adapter~\cite{chen2022vit_adapter} utilizes a CNN-based adapter which embeds local semantic features into pretrained ViTs. MIMDET~\cite{Fang2022UnleashingVV} employs a compact CNN encoder before the patch embedding layer of ViT, creating a CNN-ViT hybrid feature extractor for object detection. 

Such hybrid models are particularly well-suited for visuo-motor control applications.
This approach naturally integrates spatial locality and translation equivariance biases which lack in ViTs, but are essential for visuo-motor control.
This eliminates the need for retraining with visuo-motor specific datasets for obtaining ViTs tailored towards visuo-motor control.
However, despite active exploration in computer vision, the application of such convolutional bias integrated pretrained ViTs for visuo-motor control has been relatively unexplored (often relying on either standard ResNet or ViT models as previously mentioned in section~\ref{subsection:2.1}). Our research aims to address this gap, exploring how pretrained ViTs can be effectively adapted for control tasks while fully leveraging their well-generalized features.

\section{Method}

Our objective is to enhance the capabilities of pretrained ViTs for visuo-motor control, by introducing control-centric inductive biases during the adaptation stage. To achieve this, we propose a simple yet effective add-on module, termed {\model}.
Inspired by ViT-Adapter~\cite{chen2022vit_adapter}, {\model} is composed of a lightweight CNN encoder and a cross-attention layer \cite{chen2021crossvit}. This design enables the effective incorporation of locality and translation equivariant features extracted from the CNN encoder into ViT's patch embeddings for improved visuo-motor task performance.
From here, we denote \textit{convolutional inductive bias} to refer to both spatial locality and translation equivariance.

The ViT architecture is briefly described in Section~\ref{subsection:vit}, and the CNN module and cross-attention mechanism in {\model} are described in Section~\ref{subsection:cnn}, \ref{subsection:crossattentionmodule}, respectively.

\begin{figure}[t]
\begin{center}
\includegraphics[width=0.49\textwidth]{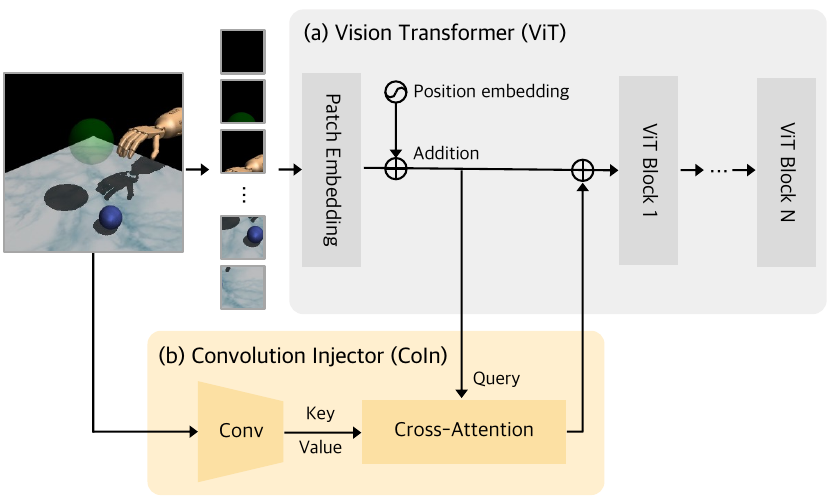}
\end{center}
\vspace{-2mm}
\caption{\textbf{Overall architecture of {\model}.} While leaving the (a) ViT architecture untouched, (b) {\model} incorporates two key modules: a CNN encoder, which captures spatial locality and translation equivariance rich features from the input image, and a cross attention module, which introduces such biases into the ViT patch token embeddings. Notably, these enhancements are seamlessly integrated without any modification to the overall ViT architecture.}
\label{fig:architecture}
\end{figure}

\subsection{Vision Transformer}
\label{subsection:vit}

In ViT, there are primarily two components: the patch embedding module and transformer encoder blocks~\cite{dosovitskiy2020image}. For an image $X \in \mathbb{R}^{H \times W \times 3}$ ($H,W$ denotes the image's resolution), the model segments the image into patches of size  $16\times16$ through the patch embedding module. Then, these patches undergoes a three-step transformation: they are (1) flattened, (2) projected into D-dimensional vectors, and (3) augmented with positional embeddings. The resultant token ($Z_0$ in Eq.~\ref{eq1}) are then fed sequentially through a series of transformer encoder blocks. 

\begin{equation}
\begin{aligned}
Z_0 &= \text{PatchEmbedding}(X),  \;\;\;\; \qquad X \in \mathbb{R}^{H \times W \times 3} \\
Z_l &= \text{Block}_l(Z_{l-1}), \;\;\;\; l = 1...L, \;\;\; Z_l \in \mathbb{R}^{N \times D}   \;\;
\end{aligned}
\label{eq1}
\end{equation}

$L$ denotes the total number of transformer encoder blocks and $N$ denotes the number of patches.

\subsection{CNN Encoder}
\label{subsection:cnn}

To adapt a standard pretrained ViT for control tasks, we introduce a lightweight CNN encoder (Figure~\ref{fig:architecture}b (left)).
Its primary role is to generate features rich in spatial locality and translation equivariance bias which will later benefit the token embeddings $Z_0$ before they proceed through the transformer encoder blocks.

The design of the CNN encoder takes inspiration from the spatial prior module described in ViT-Adapter~\cite {chen2022vit_adapter}.
Initially, it uses a standard convolutional stem~\cite {he2016resnet}, which is followed by a series of stride-2 $3 \times 3$ convolutions.

\begin{equation}
\begin{aligned}
S &=  \text{Stem} \; (X)  \qquad \qquad \; \; \; \; S \in \mathbb{R}^{\frac{H}{4} \times \frac{W}{4} \times c_0} \\
\mathcal{F}_1 &= \text{Conv}_1 \; (S) \qquad \qquad  \; \, \mathcal{F}_1  \in \mathbb{R}^{\frac{H}{8} \times \frac{W}{8} \times c_1} \\
\mathcal{F}_2 &= \text{Conv}_2 \; (\mathcal{F}_1) \qquad \qquad    \mathcal{F}_2  \in \mathbb{R}^{\frac{H}{16} \times \frac{W}{16} \times c_2} \\
\mathcal{F}_3 &= \text{Conv}_3 \; (\mathcal{F}_2) \qquad \qquad    \mathcal{F}_3  \in \mathbb{R}^{\frac{H}{32} \times \frac{W}{32} \times c_3} \\
\end{aligned}
\end{equation}

where $c_i$ denotes the hidden dimension of each layer.

As the output from the stem layer $S$ passes through subsequent $\text{Conv}_1$ to $\text{Conv}_3$,  we generate a feature pyramid consisting of multiple scales, represented as $\mathcal{F}_{\text{conv}} = \{ \mathcal{F}_1, \mathcal{F}_2, \mathcal{F}_3 \}$.
Each scale of this pyramid corresponds to a different resolution, providing a comprehensive spatial representation of the input image.
The inclusion of this multi-scale feature map array in our architecture enhances the model's ability to perceive spatial information at various resolutions, which is a key aspect for control tasks where recognizing different spatial scales is essential. Ablations on employing multi-scale feature is provided in Section~\ref{subsubsection:feature pyramids}.

Next, to makes these feature maps compatible with the ViT token embeddings, we apply $1 \times 1$ convolutions to each scale of the feature pyramid, which results in
$\mathcal{F}_1 \in \mathbb{R}^{\frac{H}{8} \times \frac{W}{8} \times D}$, $\mathcal{F}_2 \in \mathbb{R}^{\frac{H}{16} \times \frac{W}{16} \times D}$, and $\mathcal{F}_3 \in \mathbb{R}^{\frac{H}{32} \times \frac{W}{32} \times D}$, where $D$ matches the dimension of the ViT patch token embeddings $Z_0$.

\subsection{Cross Attention Module}
\label{subsection:crossattentionmodule}

To incorporate the convolutional inductive bias rich features provided by the feature pyramid $\mathcal{F}_\text{conv}$ into a pretrained ViT, we utilize the  multi-head cross attention mechanism~\citep{vaswani2017attention, alayrac2022flamingo}. 

Initially, each of the feature maps from $\mathcal{F}_\text{conv}$ is flattened and merged into a singular tensor $\mathcal{F'}_\text{conv} = \mathbb{R}^{(HW/8^2 + HW/16^2 + HW/32^2) \times D}$. 
Subsequently, we employ the output patch embeddings $Z_0$ from the ViT, as the query. $\mathcal{F}_{\text{conv}}$ is utilized both as the key and the value. 
This cross-attention mechanism enables the pretrained ViT to utilize the spatial locality and translation equivariance bias rich features extracted by the CNN encoder, which are important for downstream visuo-motor control tasks.

\begin{equation}
\begin{aligned}
\hat{Z}_0 &= Z_0 + \text{CrossAttention}(Z_0, \mathcal{F'}_\text{conv}) \\
\hat{Z}_l &= \text{Block}_l(\hat{Z}_{l-1}), \;\;\;\; l = 1 \dots L
\end{aligned}
\end{equation}
The enriched outputs $\hat{Z}_0$, which are the sum of the outputs from the cross-attention module and the original $Z_0$, are then processed through the standard encoder blocks of the original ViT transformer.

This formulation ensures a seamless and effective integration of convolutional features into the ViT architecture. It allows for a feature representation enriched with spatial priors, while simultaneously leveraging the robust and powerful representations of a pretrained ViT.

\begin{figure*}[ht]
\begin{center}
\includegraphics[width=1.0\textwidth]{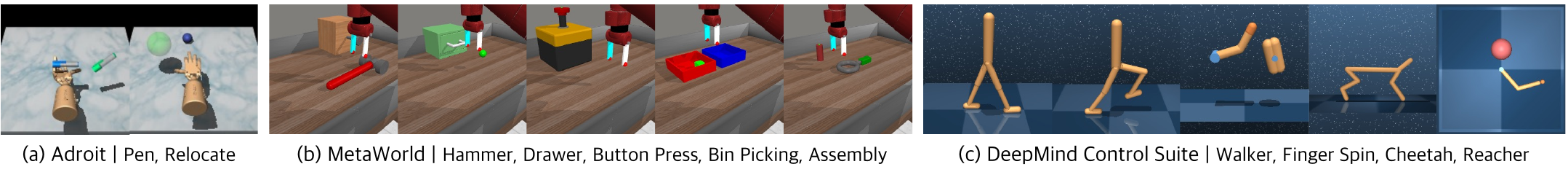}
\end{center}
\vspace{-2mm}
\caption{\textbf{Visualization of tasks used in our evaluation.} We utilize 2 tasks from Adroit, 5 tasks from Metaworld, and 5 tasks from DMC.}
\label{fig: environment}
\end{figure*}

\subsection{Implementation and Computation Requirements}
\label{subsection:computation}

We note that {\model} exhibits a significantly lower computation footprint compared to a standard ViT, thereby minimizing the additional computational burden during the finetuning stage.
To illustrate, while a standard ViT-B/16 (our primary experimental architecture) contains approximately 85.8M parameters, {\model} contains only approximately 3.1M parameters.
This amounts to merely 3.6\% of the parameter count of a ViT-B/16, highlighting {\model}'s lightweight nature.
Such a compact design makes {\model} an affordable add-on module to be additionally trained along with the ViT during the finetuning stage.
Furthermore, in the interest of computational efficiency within the cross-attention layer, we adopt a linear sparse self-attention variant~\citep{zhu2020deformableattention, chen2022vit_adapter}, which is recognized for its computational efficiency in terms of trainable parameters, training time, and memory compared to conventional global self-attention modules. Further implementation details are provided at Appendix~\ref{appendix: implementation}.

\section{Experiment Setup}

\subsection{Environments}
\label{subsection: environments}

Here we describe the environments and tasks used in our evaluation. 
We consider a total of 12 tasks across three different domains: 2 tasks from Adroit~\citep{rajeswaran2018adroit}, 5 tasks from MetaWorld~\citep{yu2020metaworld}, and 5 tasks from DMC~\citep{Tassa2018DMC}.
We provide a brief description regarding the selected tasks below (See Figure~\ref{fig: environment}).

\textbf{Adroit}~\citep{rajeswaran2018adroit} is a suite of tasks focused on dexterous manipulation.
The agent is required to control a 28-DoF anthropomorphic hand to accomplish various goal-oriented activities in a virtual 3D environment.
We focus on two challenging tasks from Adroit, 'Relocate' and 'Reorient-Pen', where the agent's objective is to either position an object at a specific target location or align it to a predetermined orientation.
These tasks serve as a measure of the robot's precision and adaptability in complex manipulation tasks.

\textbf{MetaWorld}~\citep{yu2020metaworld} requires an agent to control a Sawyer robot arm to perform various object manipulation tasks on a tabletop environment. Following prior work~\citep{majumdar2023vc1, nair2022r3m}, we utilize five tasks from MetaWorld: Assembly, Bin Picking, Button Pressing, Drawer Opening, and Hammering.

\textbf{Deepmind Control Suite (DMC)}~\citep{Tassa2018DMC} is a widely used benchmark for continuous control which involves low-level locomotion and manipulation of various difficulty. 
In our studies, we focus on five tasks from the suite: Walker Stand, Walker Walk, Reacher Easy, Finger Spin, and Cheetah Run.

\subsection{Models}

In order to validate the efficacy of our approach, we experiment on three pretrained ViT encoders which have been widely utilized as a visual encoder for control tasks.

\textbf{CLIP~}\citep{radford2021clip} is pretrained on a vast collection of web-scale image-text pairs, aligning image and language features effectively through contrastive learning.
Its capabilities extend to a variety of tasks, including manipulation and navigation~\citep{shridhar2022cliport, khandelwal2022simpleclip}.
In line with existing work, our research explores its potential as a foundational visual encoder for both manipulation and locomotion tasks.

\textbf{MVP}~\citep{radosavovic2022mvp} focuses on spatial understanding by reconstructing randomly masked patches using a massive collection of Internet and egocentric data.
MVP underlines the advantages of pretraining large visual encoders from web scale datasets for real-world robotic applications.

\textbf{VC-1}~\citep{majumdar2023vc1} seeks to extend the achievements of MVP for pretrained visual representations in robotics.
By coupling the ViT encoder and MAE pretraining objective on a more diverse dataset primarily composed of egocentric data, VC-1 attains competitive results in a wide array of visuo-motor control tasks.

\begin{table*}[t]
\caption{\textbf{Main results: {\model} with various pretrained ViTs.} Performance improvements achieved by incorporating {\model} across 12 tasks in three benchmarks (Adroit, MetaWorld, DMC) with three independent seeds. 
For each benchmark, we report the average performance and the average standard deviation of each task. {\model} is indicated in gray rows and the best results for each model are highlighted in bold.
{\model} consistently improves performance for all models and across all benchmarks.}
\vspace{2mm}
\begin{center}
\resizebox{0.84 \textwidth}{!}{
\begin{tabular}{cclcccl}
\toprule
Backbone & Model & Training Strategy & Adroit & MetaWorld  & DMC  & Mean Success  \\
\midrule \\[-2.5ex]

\multirow{4}{*}{ResNet50} & VIP &
Frozen & 58.0 \small{$\pm$ 7.6} & 92.0 \small{$\pm$ 3.5} & 64.4 \small{$\pm$ 3.8} & 71.5 \\ 
& \cite{ma2022vip} &
Finetuned & 63.3 \small{$\pm$ 4.6} & 95.5 \small{$\pm$ 3.9} & 82.4 \small{$\pm$ 1.8} & 80.4 \\ 

\cline{2-7}
\\[-2ex]

& R3M & 
Frozen & 61.3 \small{$\pm$ 6.3} & 92.5 \small{$\pm$ 2.9} & 69.8 \small{$\pm$ 3.8} & 74.5 \\ 
& \cite{nair2022r3m} & 
Finetuned & 78.7 \small{$\pm$ 3.5} & 94.9 \small{$\pm$ 3.5} & 81.8 \small{$\pm$ 1.7} & 85.1 \\ 

\midrule

\multirow{9}{*}{ViT-B} & \multirow{2}{*}{CLIP}  & 
Frozen & 38.7 \small{$\pm$ 3.2} & 60.5 \small{$\pm$ 5.1} & 37.4 \small{$\pm$ 2.3} & 45.5 \\ 
&  \multirow{2}{*}{\cite{radford2021clip}} &
Finetuned &  47.3 \small{$\pm$ 3.2} & 68.8 \small{$\pm$ 8.1} & 62.8 \small{$\pm$ 4.6}  & 59.6 \\ 
&  &
\cellcolor{gray!25}Finetuned + {\model} & \cellcolor{gray!25}\textbf{52.7 \small{$\pm$ 6.2}} & \cellcolor{gray!25}\textbf{88.8 \small{$\pm$ 3.1}} & \cellcolor{gray!25}\textbf{71.1 \small{$\pm$ 3.7}} & \cellcolor{gray!25}\textbf{70.9} {\color{red}{(+11.3)}} \\ 

\cline{2-7}
\\[-2ex]

& \multirow{2}{*}{MVP} &
Frozen & 58.0 \small{$\pm$ 3.5} & 89.6 \small{$\pm$ 5.0} & 64.6 \small{$\pm$ 5.2} & 70.7 \\ 
& \multirow{2}{*}{\cite{radosavovic2022mvp}} & 
Finetuned & 82.0 \small{$\pm$ 5.3} & 94.1 \small{$\pm$ 4.9} & 77.4 \small{$\pm$ 1.9} & 84.5 \\ 
&   & 
\cellcolor{gray!25}Finetuned + {\model}& \cellcolor{gray!25}\textbf{83.3 \small{$\pm$ 4.6}} & \cellcolor{gray!25}\textbf{94.9 \small{$\pm$ 3.5}} & \cellcolor{gray!25}\textbf{80.5 \small{$\pm$ 2.5}} & \cellcolor{gray!25}\textbf{86.2} {\color{red}{(+1.7)}} \\ 

\cline{2-7}
\\[-2ex]

& \multirow{2}{*}{VC-1} & 
Frozen & 50.0 \small{$\pm$ 5.4} & 86.7 \small{$\pm$ 5.4} & 61.0 \small{$\pm$ 3.2} & 65.9 \\ 
& \multirow{2}{*}{\cite{majumdar2023vc1}} & 
Finetuned & 73.3 \small{$\pm$ 5.2} & 93.9 \small{$\pm$ 4.0} & 74.9 \small{$\pm$ 3.5} & 80.7 \\ 
& & 
\cellcolor{gray!25}Finetuned+ {\model}   & \cellcolor{gray!25}\textbf{77.3 \small{$\pm$ 5.1}} & \cellcolor{gray!25}\textbf{95.7 \small{$\pm$ 2.2}} & \cellcolor{gray!25}\textbf{80.7 \small{$\pm$ 4.2}} & \cellcolor{gray!25}\textbf{84.6} {\color{red}{(+3.9)}} \\ 

\bottomrule 
\end{tabular}}
\end{center}
\label{table: main result}
\vskip -0.1in
\end{table*}

\subsection{Downstream Evaluation}

In this paper, we focus on adapting pretrained visual representations for visuo-motor control tasks using behavior cloning (BC) with minimal expert trajectory data to effectively learn a control policy network $\pi(\cdot)$. The objective function is defined as:
\begin{equation}
L = \sum^{N}_{i=1}\sum^{H}_{t=1}||a^i_t - \pi([z^i_t, p^i_t])||^2_2
\end{equation}

where $a_t$, $z_t$, and $p_t$ denote the expert action, the encoded visual representation, and the proprioceptive information for trajectory $i$ at timestep $t$, respectively.

Observations in the expert trajectory data consist of $256 \times 256$ RGB images, which are center-cropped to $224 \times 224$.
For ViT models, the \texttt{[CLS]} token is used as the encoded visual observation feature input to the control policy network $\pi(\cdot)$, whereas for ResNet models, the final feature map after global average pooling serves as the encoded visual observation feature input.
The default architecture utilized throughout all experiments is ViT-B/16 for ViT models and ResNet50 for ResNet models, unless otherwise specified.

For Adroit and MetaWorld tasks, agents receive proprioceptive data, which are concatenated to the encoded visual observation features before being fed into the control policy network $\pi(\cdot)$. In contrast, for DMC tasks, proprioceptive data is not available, so only the encoded visual observations are fed into $\pi(\cdot)$.

Following existing work \citep{hansen2022lfs, parisi2022unsurprising, majumdar2023vc1, nair2022r3m}, we utilize 100 expert demonstrations for Adroit and DMC, and 25 for MetaWorld, across a training span of 100 epochs.
The visuo-motor control policy's performance is evaluated every 5 epochs, with the best success rate achieved during training reported across three independent runs for each task.
For Adroit and MetaWorld, success rate serves as the primary metric, while normalized episode return is used for DMC. 
Further implementation details are provided in Appendix~\ref{appendix: implementation}.

\section{Experiments}
\label{section:experimental results}

\subsection{Main Results}
\label{subsection:rq1}

\noindent
\textbf{E2E finetuning works.}
Previous studies have primarily evaluated the efficacy of pretrained visual representation for visuo-motor control tasks by freezing the visual encoders and finetuning only the control policy network, leaving end-to-end finetuning as an open question for future investigation~\cite{parisi2022unsurprising,radosavovic2022mvp,nair2022r3m,ma2022vip}. 
End-to-end finetuning within this domain hasn't always matched the success observed in other fields such as computer vision, occasionally resulting in suboptimal performance in visuo-motor control tasks.
Previous research often suspect overfitting as a critical issue, suggesting the need of unique adaptation strategies such as performing further self-supervised pretraining on demonstration data to address these challenges~\cite{yuan2022pieg,majumdar2023vc1}.
Our findings, detailed in Appendix~\ref{appendix: fine tuning strategy}, demonstrate that applying standard optimization strategies from the computer vision domain, such as weight decay and cosine learning rate scheduling~\cite{he2022MAE}, significantly improves finetuning performance for large visual encoders in visuo-motor control, leading to enhanced task performance across all models and tasks (Table~\ref{table: main result}).

We hope this finding will encourage future research to explore and validate the adoption of computer vision finetuning practices and hyperparameters for thorough evaluation of pretrained visual encoders in visuo-motor control tasks.

\begin{figure*}[t]
\begin{center}
\includegraphics[width=1\textwidth]{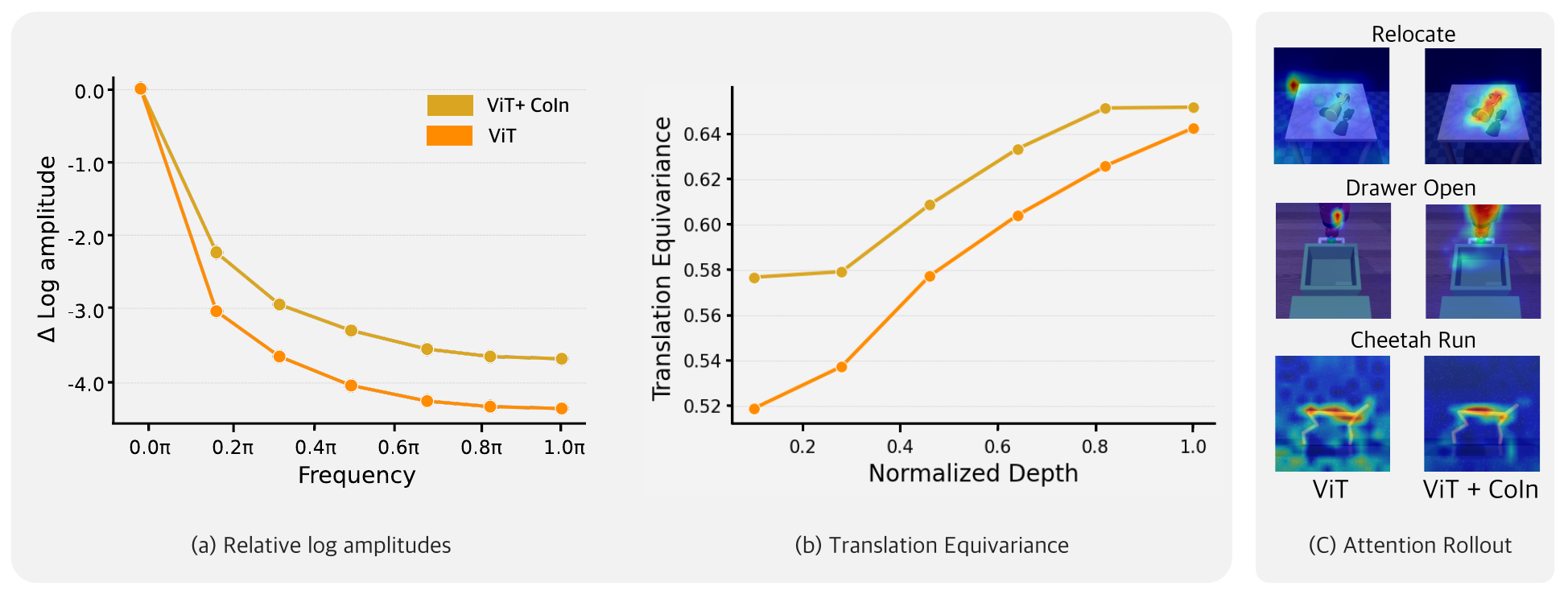}
\end{center}
\vspace{-2mm}
\caption{\textbf{(a) Comparison of the relative log amplitudes of Fourier-transformed feature maps.} ViT + {\model} incorporates beneficial inductive biases extracted from convolutional networks, allowing it to capture more high-frequency signals compared to ViT. \textbf{(b) Translation equivariance comparison.} ViT + {\model} enhances translation equivariance across intermediate representations within the ViT. 
\textbf{(c) Visualization of self-attention maps obtained through Attention Rollout.} ViT + {\model} exhibits improved focus on critical regions for visuo-motor control. 
All analysis were performed on VC-1 and averaged across all 12 tasks.
}
\label{fig:fourier_rollout}
\end{figure*}

\noindent
\textbf{Effectiveness of {\model}.} 
The integration of {\model} with various pretrained ViTs leads to notable performance enhancements across all baseline ViT models and their associated control tasks, as detailed in Table~\ref{table: main result}. Specifically, when {\model} is combined with CLIP, there is a significant increase in the mean performance by 11.3 points. Furthermore, the addition of {\model} also benefits MVP and VC-1, boosting their mean performance by 1.7 and 3.9 points, respectively.

The unique efficacy of {\model} with CLIP, as compared to its integration with MVP and VC-1, can be ascribed to the distinct nature of CLIP's pretraining datasets. MVP and VC-1 are pretrained on datasets with an egocentric perspective, making them naturally compatible with environments such as Adroit and MetaWorld, which require egocentric visual inputs from robot agents. 
Conversely, CLIP, which is pretrained on diverse web-scale image-text pairs, does not initially possess these egocentric, control-centric features. By integrating {\model}, CLIP is endowed with control-oriented inductive biases, allowing for a significant enhancement in its ability to adapt features for motor control tasks. This demonstrates that {\model} can be especially beneficial for pretrained ViT models which lack control-centric visual features.
We would also like to note that despite being pretrained with egocentric data, MVP and VC-1 still lack control-specific inductive biases necessary for certain tasks, as evidenced by the performance gains when incorporating {\model}.

Further comparison with ResNet-based pretrained visual representation methods reveals an intriguing aspect of {\model}'s performance.
Initially, R3M outperforms ViTs, indicating the advantage of inductive biases from convolution layers for these tasks. However, {\model}'s integration significantly enhances ViT performance, allowing them to meet or even surpass R3M's success rates in cases like MVP. This demonstrates {\model}'s effectiveness in adapting ViTs for control tasks, especially when the pretrained models lack control-oriented features. Full performance table is available in Appendix~\ref{table: full_main_result}.

\subsection{Analysis of {\model} Visual Features}
\label{subsection: analysis}

In this section, we perform an in-depth analysis regarding the distinct properties of the visual features learned by ViTs, both with and without the incorporation of {\model}.
Our primary objective is to determine whether incorporating {\model} in a standard ViT does indeed induce convolutional inductive bias rich visual features.
We utilize VC-1 as our baseline model throughout our analysis experiments in this section.

\textbf{Capturing high frequency.} 
CNNs excel in detecting detailed, high-frequency elements in images, including local details such as texture, edges and contours~\cite{bai2022improving}, which is an essential property for effective visuo-motor control.
Thus, we assess whether {\model} helps ViTs in capturing such valuable high-frequency elements by examining the relative log amplitudes in the Fourier transformed feature maps~\citep{chen2022vit_adapter, si2022iformer, park2022howvit}.
Figure~\ref{fig:fourier_rollout}a illustrates our findings: finetuning ViTs with {\model} significantly improves its ability to detect high-frequency signals compared to a standard ViT.
This highlights the crucial role of {\model} in effectively instilling spatial inductive biases to ViTs, thereby enhancing the performance of ViTs for downstream control tasks.

\textbf{Translation equivariance.} To evaluate whether the convolutional characteristics of {\model} enhance the learning of translation equivariant features for pretrained ViTs, we conducted a synthetic experiment following~\citet{bruintjes2023vitaffects}. 
Specifically, we computed the Pearson correlation between $f_{1:i}(T(X))$ and $T(f_{1:i}(X))$ for all $i=1,2,...,N$, where $T$ represents translations (diagonal shifts), $f_i$ denotes the i-th intermediate layers of the ViT (i.e., $f_{1:i}(X) = f_i \cdot f_{i-1} \cdot ... \cdot f_1(x)$) and N refers to the total number of layers.
Higher correlation values indicate that the model has learned higher translation equivariance.

While position embeddings in ViTs are known to present challenges for learning translation equivariance~\cite{xuequivariance20232,dai2021coatnet,dingequivariance2023reviving}, {\model} alleviates this issue by directly injecting translation equivariance rich features into the output patches of the patch-embedding layer (Figure~\ref{fig:architecture}).
As illustrated in Figure~\ref{fig:fourier_rollout}b, we observed that incorporating {\model} with ViTs enhances translation equivariance throughout the ViT intermediate representations.

\textbf{Attention visualization.} Additionally, we employ Attention Rollout~\cite{abnar2020attentionrollout, vitattentionrollout} to visualize the self-attention maps of both VC-1 with and without {\model} (Figure~\ref{fig:fourier_rollout}c).
This qualitative analysis further demonstrates that when equipped with {\model}, ViTs effectively focus more on image regions that are semantically relevant for visuo-motor control.
This also highlights the efficacy in further providing ViTs with spatial locality and translation equivariance rich features via {\model}.
More qualitative results can be found in Appendix~\ref{appendix: visualization}.

\subsection{Comparison with Adapters}
\label{subsection: peft}

In this section, we aim to address a fundamental question: \textit{Does the performance improvement of {\model} stem primarily from the utilization of additional parameters?}
To assess this, we compare {\model} with two well established adapter-based methods, RoboAdapter~\citep{sharma2023roboadapter} and Adaptformer~\citep{chen2022adaptformer}.
Although these adapter-based methods originally focus on parameter-efficient finetuning (PEFT), where the pretrained visual encoder is frozen and only the lightweight additional modules are finetuned for task adaptation~\citep{houlsby2019adapter, hu2021lora}, we explore a full finetuning variant of this approach where the pretrained visual encoder is finetuned alongside with the additional adapter modules. 
Such full finetuning variant provides an efficient means in incorporating additional task-specific parameters during finetuning.

\begin{table}[h]

\caption{\textbf{Full finetuning performance against adapter methods.} 
We report the mean performance of full finetuning across all 12 tasks in Adroit, MetaWorld, and DMC. Evaluations were conducted using CLIP and VC-1 with ViT-B. Underscored values indicate the hidden dimension size of the adapter modules. For detailed results, refer to Table~\ref{table: fullfinetune_appendix}.}
\begin{center}
\resizebox{0.4\textwidth}{!}{
\begin{tabular}{lccc}
\toprule
Model & Module & \# trainable params  &  Mean \\
\midrule \\[-2.5ex]

\multirow{5}{*}{CLIP} & X & 85.8M & 59.6 \\ 

& AdaptFormer$_{64}$ & +1.2M & 59.4 \\ 
& RoboAdapter$_{64}$ & +1.2M & 59.2 \\
& RoboAdapter$_{192}$ & +3.5M & 59.1 \\
& \cellcolor{gray!25}{\model} & \cellcolor{gray!25}+3.1M & \cellcolor{gray!25}\textbf{70.9}\\  

\\[-2ex]
\cline{1-4}
\\[-2ex]

\multirow{5}{*}{VC-1} & X & 85.8M & 80.7 \\ 

& AdaptFormer$_{64}$ & +1.2M & 82.3 \\ 
& RoboAdapter$_{64}$ & +1.2M & 83.0 \\
& RoboAdapter$_{192}$ & +3.5M & 82.5 \\
& \cellcolor{gray!25}{\model} & \cellcolor{gray!25}+3.1M & \cellcolor{gray!25}\textbf{84.6}\\  

\bottomrule 
\end{tabular}}
\end{center}
\label{table: finetune_peft}
\vskip -0.1in
\end{table}

Our findings, as detailed in Table~\ref{table: finetune_peft}, demonstrate that none of the adapter-based baseline methods match the performance of {\model}.
Notably, for CLIP, only {\model} was able to enhance CLIP's performance while all other adapter baselines failed to provide any performance gains.
This observation aligns with our explanation of why {\model} offers greater performance gains for CLIP compared to VC-1.
The lack of performance gains from other adapter baselines can be attributed to their composition, which consists solely of MLPs and does not include any control-oriented inductive bias.
As a result, these baselines are ineffective in helping CLIP learn control-oriented features.
In contrast, {\model}'s ability to impart control-oriented inductive biases significantly enhances CLIP's capacity to adapt features for motor control tasks.

This finding suggests that the effectiveness of {\model} is not merely due to the inclusion of additional trainable parameters.
Instead, it significantly stems from the strategic integration of locality and translation equivariance biases, which are lacking in standard ViTs and are particularly beneficial for pretrained models lacking control-centric visual features.

\begin{table}[h]

\caption{\textbf{PEFT performance against adapter methods.} 
We report the mean performance of parameter-efficient finetuning across all 12 tasks in Adroit, MetaWorld, and DMC. Evaluations were conducted using CLIP and VC-1 with ViT-B. Underscored values indicate the hidden dimension size of the adapter modules. For full results, refer to Table~\ref{table: peft_appendix}.}
\begin{center}
\resizebox{0.4\textwidth}{!}{
\begin{tabular}{lccc}
\toprule
Model & Module & \# trainable params  &  Mean \\
\midrule \\[-2.5ex]

\multirow{5}{*}{CLIP} & X & -- & 45.5 \\ 

& AdaptFormer$_{64}$ & 1.2M & 51.3 \\ 
& RoboAdapter$_{64}$ & 1.2M & 50.1 \\
& RoboAdapter$_{192}$ & 3.6M & 50.3 \\
& \cellcolor{gray!25}{\model} & \cellcolor{gray!25}3.3M & \cellcolor{gray!25}\textbf{67.9}\\  

\\[-2ex]
\cline{1-4}
\\[-2ex]

\multirow{5}{*}{VC-1} & X & -- & 65.9 \\ 

& AdaptFormer$_{64}$ & 1.2M & 78.8 \\ 
& RoboAdapter$_{64}$ & 1.2M & 78.8 \\
& RoboAdapter$_{192}$ & 3.6M & 78.9 \\
& \cellcolor{gray!25}{\model} & \cellcolor{gray!25}3.3M & \cellcolor{gray!25}\textbf{79.5}\\  

\bottomrule 
\end{tabular}}
\end{center}
\label{table: frozen_peft}
\vskip -0.1in
\end{table}

Additionally, we also conduct experiments under the PEFT scenario, which is the typical approach for training adapter-based methods.
In this approach, the visual encoder is frozen and only the lightweight additional modules are finetuned.
To achieve this, {\model} was minimally modified to include additional lightweight bottleneck MLP layers (merely 0.2M additional parameters) for the first two ViT encoder block.
These layers are designed to process novel patch embeddings enriched with convolutional inductive biases, which the frozen pretrained ViT had not previously encountered.

Despite {\model} not being originally designed for parameter-efficient transfer, the results in Table~\ref{table: frozen_peft} demonstrate {\model}'s superior performance against traditional adapter methods.
This highlights the potential of our approach for parameter efficient transfer learning of visual encoders in control tasks.
In addition, similar to the full finetuning scenario, the apparent effectiveness of {\model} over other adapter baselines for CLIP further emphasizes the effectiveness of {\model}'s ability to inject control-centric inductive biases crucial for visuo-motor tasks into pretrained ViTs which lack such biases,

\subsection{Ablation Study}
\label{subsection:ablation}

\subsubsection{Feature Pyramid Components}
\label{subsubsection:feature pyramids}

In Section~\ref{subsection:cnn}, we discussed the rationale behind incorporating multi-scale feature maps in {\model} instead of relying solely on a single feature map from its convolutional module. The central idea is that multi-scale feature maps can significantly improve the representational capabilities of visuo-motor control policies by capturing objects at different scales. This multi-scale approach is deemed critical for visuo-motor tasks, where the size and appearance of objects or obstacles can greatly vary. To evaluate our hypothesis, we evaluated a variant of {\model} that exclusively uses the $\mathcal{F}_2$ feature map. This feature map matches the resolution of ViT-B/16, meaning it is $\frac{1}{16}$th the size of the input resolution and does not offer a variety of feature scales.

\begin{table}[ht]
\caption{\textbf{Feature Pyramid Comparison.}
Compared to using a single-scale feature map $\{ \mathcal{F}_2\}$, utilizing multi-scale feature map $\{\mathcal{F}_1, \mathcal{F}_2, \mathcal{F}_3\}$ is beneficial for {\model}. 
}
\begin{center}
\resizebox{0.48 \textwidth}{!}{
\begin{tabular}{lccccc}
\toprule
Model \& & \multirow{2}{*}{Feature Pyramid} & \multirow{2}{*}{Adroit} & Meta- & \multirow{2}{*}{DMC} & \multirow{2}{*}{Mean}  \\
Strategy &   & & World & & \\
\midrule \\[-2.5ex]

\multirow{2}{*}{CLIP +}
& X & 47.3 \small{$\pm$ 3.2} & 68.8 \small{$\pm$ 8.1} & 62.8 \small{$\pm$ 4.6} & 59.6 \\

\multirow{2}{*}{Finetuned}
& $\{ \mathcal{F}_2 \}$
& 51.3 \small{$\pm$ 5.2} & 86.1 \small{$\pm$ 4.0} &  73.9 \small{$\pm$ 3.3} & 70.4\\

& \cellcolor{gray!25}$\{ \mathcal{F}_1, \mathcal{F}_2, \mathcal{F}_3 \}$
& \cellcolor{gray!25}52.7 \small{$\pm$ 6.2} & \cellcolor{gray!25}88.8 \small{$\pm$ 3.1} & \cellcolor{gray!25}71.1 \small{$\pm$ 3.7} & \cellcolor{gray!25}\textbf{70.9}\\

\\[-2ex]
\cline{1-6}
\\[-2ex]

\multirow{2}{*}{VC-1 +}
& X & 73.3 \small{$\pm$ 5.2} & 93.9 \small{$\pm$ 4.0} & 74.9 \small{$\pm$ 3.5} & 80.7 \\

\multirow{2}{*}{Finetuned}
& $\{ \mathcal{F}_2 \}$
& 76.0 \small{$\pm$ 3.5} & 95.2 \small{$\pm$ 3.4} &  79.9 \small{$\pm$ 3.1} & 83.7\\

& \cellcolor{gray!25}$\{ \mathcal{F}_1, \mathcal{F}_2, \mathcal{F}_3 \}$
& \cellcolor{gray!25}77.3 \small{$\pm$ 5.1} & \cellcolor{gray!25}95.7 \small{$\pm$ 2.2} & \cellcolor{gray!25}80.7 \small{$\pm$ 4.2} & \cellcolor{gray!25}\textbf{84.6}\\

\bottomrule 
\end{tabular}}
\label{table:singlescale}
\end{center}
\vskip -0.1in
\end{table}

Table~\ref{table:singlescale} indicates that even the single-scale variants significantly improves the performance against the vanilla baseline, clearly highlighting the benefits of integrating inductive biases tailored to control tasks. Additionally, incorporating multi-scale feature maps yielded even better results, demonstrating the efficacy of providing varied scale perspectives.

We note that the integration of multi-scale feature maps into {\model} is both efficient and cost-effective, as they are intrinsically generated from the lightweight convolutional layers.
Given these empirical observations, we set the multi-scale feature pyramid as our standard setup for {\model}.

\subsubsection{Scaling Study}

To further understand the effectiveness of {\model} on larger ViT models, we performed experiments where {\model} is applied to ViT-L/14 across all 12 tasks considered in this work.
Table~\ref{table: scaling} details the results of finetuning VC-1 with and without {\model}, for both ViT-B/16 and ViT-L/14. For performance results of each benchmark, refer to Table~\ref{table: scaling_full_table}.

\begin{table}[ht]
\caption{\textbf{{\model} with larger scale.} {\model} boosts visuo-motor control task performance for both scales, where ViT-B + {\model} outperforms ViT-L with significantly fewer parameters. }
\begin{center}
\resizebox{0.43  \textwidth}{!}{
\begin{tabular}{lccc}
\toprule
Model \& & Backbone Scale \& & \# trainable &  \multirow{2}{*}{Mean}  \\
Strategy & Additional Module & params &  \\
\midrule \\[-2.5ex]

\multirow{3}{*}{VC-1+} & ViT-B & 85.8M & 80.7 \\ 
 \multirow{3}{*}{Finetuned} & \cellcolor{gray!25}ViT-B + {\model} & \cellcolor{gray!25}88.9M &   \cellcolor{gray!25}84.6 \\
\cline{2-4}
\\[-1.7ex]
& ViT-L  & 303.3M & 83.4 \\
 & \cellcolor{gray!25}ViT-L + {\model} & \cellcolor{gray!25}307.7M & \cellcolor{gray!25}84.5 \\

\bottomrule 
\end{tabular}}
\end{center}
\label{table: scaling}
\vskip -0.1in
\end{table}

The findings in Table~\ref{table: scaling} lay out three important aspects.
First, finetuning larger pretrained encoders, as expected, yield better performance than smaller pretrained encoders in downstream visuo-motor control tasks.
Second, the benefits of injecting convolutional inductive bias via {\model} works in tandem with increasing model size, enhancing performance regardless of the model's scale.
Lastly, a particular surprising observation is that when ViT-B is paired with {\model}, despite having significantly fewer parameters, it outperforms ViT-L. 
This emphasizes the role of {\model} in enhancing smaller encoders to reach the efficacy levels of their larger counterparts.

Overall, these findings underscore the scalability and efficacy of {\model} in enhancing ViT models for complex visuo-motor control tasks, revealing its potential for application in larger encoders.

\section{Discussion and Conclusion}
\label{section: conclusion}

In our study, we explore the challenges encountered by pretrained ViTs when applied to visuo-motor tasks.
Particularly, while the weak inductive bias of ViTs are advantageous for large-scale pretraining, it limits their applicability in control-specific scenarios. 
To address this, we introduce a simple lightweight module, {\model}, designed to inject ViTs with convolutional inductive biases which are beneficial in performing effective visuo-control.
This allows pretrained ViTs to leverage both their strong visual representations and beneficial biases for downstream visuo-motor control tasks provided by {\model}. Our thorough evaluation across a variety of visuo-motor control tasks confirms the consistent advantages and efficiency of {\model}.

In this work, we primarily focus on learning policies using behavior cloning, to highlight how {\model} effectively enhances ViTs for control tasks under limited data availability. We believe that extending {\model} to reinforcement learning for complex robotic tasks is a valuable avenue for future investigation.
Moreover, while our current experiments are conducted within simulated environments, real-world robot experiments may present additional challenges and leave the evaluation of {\model} on real-world hardware as future work.

\section*{Acknowledgements}

This work was supported by Institute for Information \& communications Technology Promotion (IITP) grant funded by the Korea government (MSIT) (No.RS-2019-II190075 Artificial Intelligence Graduate School Program(KAIST), RS-2021-II212068 Artificial Intelligence Innovation Hub) and the National Research Foundation of Korea (NRF) grant funded by the Korea government (MSIT) (No. NRF-2022R1A2B5B02001913).

\section*{Impact Statement}

Our add-on module, {\model}, which injects convolutional inductive biases into pretrained Vision Transformers (ViTs), enables efficient adaptation of pretrained ViTs for visuo-motor control tasks. We anticipate that as foundation models such as ViTs continue to advance in the computer vision field, the performance of visuo-motor control systems will improve in tandem. Moreover, we acknowledge the potential risks associated with the current rapid advancements and potential misuse of such technologies. However, we believe that there are no specific ethical considerations in this paper which we feel must be specifically highlighted here.

\bibliography{main}
\bibliographystyle{icml2024}

\newpage
\appendix
\onecolumn

\section{Finetuning strategy}
\label{appendix: fine tuning strategy}

Prior research mainly evaluate pretrained visual encoder for visuo-motor control tasks by keeping them frozen and finetuning only the control policy.
However, we argue that this approach does not fully assess the capabilities of pretrained visual encoders for visuo-motor control tasks. 
To understand their effectiveness, it is crucial to examine the effectiveness of pretrained visual encoders both when they are frozen and when they are finetuned.
For instance, previous studies have suggested that linear probing may not accurately correlate with transfer learning performance~\citep{chen2021simsiam}, with some findings showing inconsistent rankings across tasks (See Figure 9 in \citep{he2022MAE})

Despite this, how to optimize pretrained visual encoder for visuo-motor control tasks is an under-researched question which has not received the attention it should by the research community. 
Addressing this gap, our experiments demonstrate that applying ViT finetuning strategies commonly utilized by the computer vision community are also effective for visuo-motor control tasks. 

\begin{table}[h]
\vspace{-3mm}
\caption{\textbf{Finetuning strategy ablation}. Here we present the mean succes of VC-1 with ViT-B finetuned across all 12 tasks in Adroit, Metaworld, and DMC. The grey row indicates our default setup for finetuning ViTs. Cosine LR indicates cosine learning rate decay and LLDR indicates layer-wise learning rate decay.}
\begin{center}
\resizebox{0.5 \textwidth}{!}{
\begin{tabular}{l ccccc}
\toprule
& Finetune & Weight decay & Cosine LR & LLDR & Mean Success $\uparrow$ \\
\midrule \\[-2.4ex]
(a) &  -           &  -         & -           &-           & 65.9    \\
(b) &  \checkmark  &  -         & -           &-           & 55.1    \\
(c) &  \checkmark  & \checkmark & -           &-           & 58.9    \\
(d) &  \checkmark  & \checkmark & \checkmark  &-           & 76.5    \\
\rowcolor{shadecolor}                                 
(e) &  \checkmark  & \checkmark & \checkmark  & \checkmark & 80.7    \\

\bottomrule
\end{tabular}}
\label{table:finetuning strategy}
\end{center}
\end{table}

Our results, as shown in Table~\ref{table:finetuning strategy} (a) and (b), first reveal an initially counter-intuitive finding which has often been observed by previous research~\citep{yuan2022pieg, majumdar2023vc1}: simply finetuning VC-1 with ViT-B results in deteriorated performance compared to its frozen counterpart.
This is counter intuitive since further training the visual encoder on in-domain specific data should increase performance, not deteriorate the performance of the control policy.
We speculate that the main cause of this performance deterioration is due to overfitting.
By implementing a combination of weight decay, cosine learning rate decay, and layer-wise learning rate decay~\citep{bao2021beit, clark2020electra}, which are techniques commonly used in finetuning ViTs~\citep{he2022MAE, bao2021beit, dong2022clipfinetune, steiner2021trainvit}, we observe significant performance improvements.
Specifically, applying weight decay led to a 3.8 points increase ($55.1 \rightarrow 58.9$), cosine learning rate decay resulted in a further 17.6 point boost ($58.9 \rightarrow 76.5$), and layer-wise learning rate decay~\citep{clark2020electra, bao2021beit} added an additional performance improvement of 4.2 points ($76.5 \rightarrow 80.7$).
These strategies demonstrate the potential of finetuning to unlock the full capabilities of pretrained visual encoders for visuo-motor control tasks.

\section{Comparison with ViT-Adapter}
Here, we describe in detail how {\model} differs compared to the context of existing computer vision literature which introduce convolutional inductive biases to the ViT architecture, particularly in relation to ViT-Adapter~\cite{chen2022vit_adapter}. We provide a detailed explanation to elucidate the difference between {\model} and ViT-Adapter and illustrate the advancements that {\model} offers through experimental results.

We would first like to clarify that {\model} and ViT-Adapter differ in (i) motivation and (ii) practical implementation.
The main motivation behind {\model} is in \textit{injecting convolutional inductive biases into pretrained ViTs}, as spatial locality and translation equivariance are beneficial properties in performing precise visuo-motor control.
In contrast, the principal motivation behind ViT-Adapter is to construct an effective multi-scale feature map from pretrained ViT representations. This multi-scale feature map is subsequently used by dense prediction heads~\cite{He2017_maskrcnn, cai2019_cascade, Kirillov2019_semanticfpn} to perform tasks such as object detection and semantic segmentation.
Although {\model} incorporates elements inspired by ViT-Adapter, this difference in their motivations results in a key distinction: {\model} does not require the Extractor module present in ViT-Adapter and operates exclusively with the Injector module.

In ViT-Adapter, the Extractor module constructs effective multi-scale convolutional feature maps $\mathcal{F}_{\text{conv}} = \{ \mathcal{F}_1, \mathcal{F}_2, \mathcal{F}_3 \}$ by distilling representations from a pretrained ViT. This feature map serves as the output of ViT-Adapter.
Conversely, {\model} focuses on injecting convolutional inductive bias rich features into pretrained ViTs for effective visuo-motor control. 
Consequently, the Injector module is essential for {\model} while the Extractor module is unnecessary from a motivation standpoint. In addition, {\model} employs the ViT \texttt{[CLS]} token, rather than the multi-scale feature maps $\mathcal{F}_{\text{conv}} = \{ \mathcal{F}_1, \mathcal{F}_2, \mathcal{F}_3 \}$ for predicting actions. This further underscores the redundancy of the Extractor module in terms of practical implementation.
Therefore, driven by its core motivation and implementation strategy, {\model} solely utilizes the Injector module.

\begin{table}[h]
\caption{\textbf{Performance comparison between {\model} and ViT-Adapter}. 
We report the mean performance results of {\model} against ViT-Adapter across all 12 tasks in three benchmarks (Adroit, MetaWorld, DMC) with three independent seeds. {\model} significantly outperforms ViT-Adapter in terms of computational cost, inference speed, and mean performance.}
\begin{center}
\resizebox{0.83\textwidth}{!}{
\begin{tabular}{lcccccc}
\toprule
Model (ViT-B/16) & Additional component & \# trainable params & MACs & Inference speed &  Mean \\
\midrule \\[-2.5ex]

VC-1 & X & 85.8M &  16.88G & 6.04 ms & 80.7\\ 

VC-1 + ViT-Adapter & Injector \& Extractor & 103.2M &  26.07G & 14.50 ms & 80.8\\ 
VC-1 + {\model} (Ours) & \cellcolor{gray!25}Injector only & \cellcolor{gray!25}88.9M &  \cellcolor{gray!25}19.06G & \cellcolor{gray!25}8.78 ms & \cellcolor{gray!25}84.6 \\

\bottomrule 
\end{tabular}}
\end{center}
\label{table: vitadapter_appendix}
\end{table}

Moreover, through extensive experiments across 12 varied visuo-motor control tasks, we empirically observed that {\model} outperforms ViT-Adapter significantly in (i) computational cost (26.07G → 19.06G), (ii) inference speed (14.50 ms → 8.78 ms) and in (iii) mean score performance (80.8 → 84.6). These empirical results clearly supports our architectural choices behind {\model} and demonstrates the advantages of {\model} over ViT-Adapter for visuo-motor control tasks. We note that computational costs were calculated using ptflops\footnote{Code : \url{https://github.com/sovrasov/flops-counter.pytorch}}~\cite{ptflops} and inference speed was calculated on a single RTX-3090 GPU using a single input image with a resolution of 224 x 224.

In summary, {\model}'s contribution is in its optimized and practical architecture tailored towards effectively adapting large-scaled pretrained ViTs for visuo-motor control applications. {\model} differs from ViT-Adapter in terms of both motivation and architectural design, while also demonstrating superior performance in computational cost, inference speed and mean performance.

\section{Implementation Details}
\label{appendix: implementation}

\subsection{Visual encoder}

Detailed hyperparameters for finetuning pretrained visual encoders with and without {\model} are listed in Table~\ref{table:common_hyperparameters}. The same set of hyperparameters is shared across all pretrained visual encoders, regardless of the task and architecture, with the following exceptions: (i) a smaller learning weight is used for CLIP-related experiments, and (ii) different layer-wise learning rate decay values are applied between ViT-based models and ResNet-based models.

\begin{table}[ht]
\caption{\textbf{Visual encoder finetuning hyperparameters.} The same set of hyperparameters is applied across all pretrained ViT and ResNet methods - except for (i) using a smaller learning weight for CLIP-related experiments and (ii) applying different layer-wise learning rate decay values between ViT-based models and ResNet-based models - when performing finetuning regardless of with or without {\model} for all tasks.}

\vskip 0.15in
\begin{center}
\begin{tabular}{l c}
\toprule
\textbf{Hyperparameter}                 & Value              \\
\hline \\[-2.0ex]
Optimizer                               &AdamW                \\
\multirow{2}{*}{Learning rate}          &CLIP : $1\times 10^{-4}$ \\
                                        &others : $1\times 10^{-3}$ \\
Optimizer momentum                      &$\beta_1,\beta_2 = 0.9, 0.999$ \\
\multirow{2}{*}{Layer-wise lr decay}    &ViT : 0.75 \\
                                        &ResNet : 1.0 \\
Weight decay                            &0.05   \\
Batch size                              &256                \\
Learning rate schedule                  &cosine decay           \\
Warmup epochs                           &5                  \\
Training epochs                         &100  \\
\bottomrule 
\end{tabular}
\end{center}
\label{table:common_hyperparameters}
\end{table}

\subsection{Control policy}

We closely follow the architecture and training hyperparameters of the control policy network from prior work~\cite{majumdar2023vc1, hansen2022lfs}.
Specifically, the control policy network is a 4-layer MLP with 256 hidden units each and ReLU activation. Additionally, the control policy includes a 1D BatchNorm layer at the beginning to normalize the pretrained visual representations.
As in VC-1~\cite{majumdar2023vc1}, we also use frame-stacking, where the visual encoder individually encodes each observation in the stack of recent observations. The control policy then fuses the encoded features using Flare~\cite{shang2021flare}. Detailed hyperparameters for finetuning the control policy network are listed in Table~\ref{table:policy_common_hyperparameters}.
As with their visual encoder counterpart, the same set of hyperparameters are shared across all tasks regardless of the visual encoder architecture.

\begin{table}[ht]
\caption{\textbf{Control policy finetuning hyperparameters.} The same hyperparameters are applied across all tasks.}

\vskip 0.15in
\begin{center}
\begin{tabular}{l c}
\toprule
\textbf{Hyperparameter}                 & Value              \\
\hline \\[-2.0ex]
Optimizer                               &Adam                \\
Learning rate                           &$1\times 10^{-3}$   \\
Optimizer momentum                      &$\beta_1,\beta_2 = 0.9, 0.999$ \\
Hidden units                            & 256, 256, 256    \\
Frames stacked                           & 3                \\
Batch size                              &256                \\
Training epochs                         &100  \\
\bottomrule 
\end{tabular}
\end{center}
\label{table:policy_common_hyperparameters}
\end{table}

\subsection{{\model}}

\textbf{Positional embeddings}
For the multi-scale convolutional feature maps $\mathcal{F}_{\text{conv}} = \{ \mathcal{F}_1, \mathcal{F}_2, \mathcal{F}_3 \}$ within {\model}'s Injector module, we assign separate learnable 1D embeddings for each scale which act as positional embeddings.
For example, feature map $\mathcal{F}_1$ with scale $H/8 \times W/8$ is combined with a positional embedding $E_{\mathcal{F}_1} \in \mathbb{R}^d$, while feature map $\mathcal{F}_3$ with scale $H/32 \times W/32$ is combined with a different embedding $E_{\mathcal{F}_3} \in \mathbb{R}^d$.
These learnable 1D embeddings enable the model to distinguish features extracted from different scales.

\textbf{Cross-attention mechanism}
{\model}'s cross-attention mechanism uses deformable attention~\cite{zhu2020deformableattention} to address the inherent computational inefficiency of global self-attention where each query token has a global spatial receptive field and examines every key/value token when computing attention weights. 
This leads to quadratic computation requirements in terms of the total number of query and key tokens.
Motivated by the key underlying principle of deformable convolution~\cite{dai2017deformableconv}, our implementation enables each query patch token (i.e., reference points) in the deformable attention module to selectively focus on a small set of spatially relevant locations (i.e., sampling locations) by predicting a fixed number of sampling offsets respective to the reference point.
This selective attention module circumvents the necessity of computing attention weights for every key token for each query token, concentrating instead on a fixed number of key points identified for each query patch token.
As a result, this significantly reduces the computational complexity to linear terms relative to the number of query tokens, enhancing the efficiency of the cross-attention process.
Therefore, employing deformable attention aligns with our goal in achieving a scalable, fast, and effective cross-attention mechanism for {\model}, while also focusing on spatially relevant local locations.

\textbf{Architecture configurations}
For the deformable attention, we fix the number of sampling points to 4, and the number of attention heads to 12 and 16 for ViT-B and ViT-L, respectively. In addition, we downsize the feature embedding size in our Injector module to save computation overhead, where the hidden dimension size is 192 for ViT-B and 256 for ViT-L. We only use a single Injector module in all experiments, as additional Injector modules did not provide additional gains.


\newpage

\section{Main Results}

\begin{table}[h]
\begin{center}
\caption{\textbf{Success rate of each individual task and model.} We present the success rate and standard deviation for each task and model we evaluate during our experiments for Section~\ref{section:experimental results} before aggregating them for each benchmark. All tasks were evaluated with three independent seeds.}
\vspace{1.4ex}
\resizebox{0.99\textwidth}{!}{
\begin{tabular}{l|cc|ccccc|ccccc}
\toprule &\\[-2.5ex]
\multirow{2}{*}[1.5ex]{Task} & \multicolumn{2}{c|}{Adroit} & \multicolumn{5}{c|}{MetaWorld} & \multicolumn{5}{c}{DMC} \\
Model & pen & relocate & hammer & drawer open & button press & bin picking & assembly & walker stand & walker walk & reacher easy & finger spin & cheetah run
\\[0.2ex]
\midrule \\[-2.7ex]
VIP Frozen        & $  78.7$ \small{$\pm$ 2.3} & $   37.3$ \small{$\pm$ 12.9} & $   94.7$ \small{$\pm$ 4.6} & $   98.7$ \small{$\pm$ 2.3} & $   82.7$ \small{$\pm$ 2.3} & $   93.3$ \small{$\pm$ 2.3} & $   90.7$ \small{$\pm$ 6.1} & $   76.9$ \small{$\pm$ 8.0} & $   47.2$ \small{$\pm$ 1.7} & $  89.7$ \small{$\pm$ 4.8} & $   70.2$ \small{$\pm$ 0.4} & $   38.2$ \small{$\pm$ 4.3} \\

VIP Finetuned      & $  80.0$ \small{$\pm$ 0.0} & $   46.7$ \small{$\pm$ 9.2} & $   98.7$ \small{$\pm$ 2.3} & $   100.0$ \small{$\pm$ 0.0} & $   96.0$ \small{$\pm$ 4.0} & $   88.0$ \small{$\pm$ 4.0} & $   94.7$ \small{$\pm$ 9.2} & $  94.5$ \small{$\pm$ 1.3} & $   87.7$ \small{$\pm$ 1.6} & $   89.1$ \small{$\pm$ 2.6} & $   69.8$ \small{$\pm$ 0.2} & $   70.7$ \small{$\pm$ 3.4} \\

\midrule \\[-2.7ex]
R3M Frozen        & $  78.7$ \small{$\pm$ 4.6} & $   44.0$ \small{$\pm$ 8.0} & $   100.0$ \small{$\pm$ 0.0} & $   100.0$ \small{$\pm$ 0.0} & $   74.7$ \small{$\pm$ 6.1} & $   93.3$ \small{$\pm$ 2.3} & $   94.7$ \small{$\pm$ 6.1} & $  88.2$ \small{$\pm$ 1.0} & $   64.5$ \small{$\pm$ 6.8} & $   91.0$ \small{$\pm$ 6.1} & $   68.7$ \small{$\pm$ 1.0} & $   36.7$ \small{$\pm$ 3.9} \\

R3M Finetuned      & $  82.7$ \small{$\pm$ 2.3} & $   74.7$ \small{$\pm$ 4.6} & $   100.0$ \small{$\pm$ 0.0} & $   100.0$ \small{$\pm$ 0.0} & $   84.0$ \small{$\pm$ 10.6} & $   93.3$ \small{$\pm$ 2.3} & $   97.3$ \small{$\pm$ 4.6} & $  96.6$ \small{$\pm$ 1.3} & $   89.7$ \small{$\pm$ 1.0} & $   91.8$ \small{$\pm$ 4.3} & $   68.9$ \small{$\pm$ 0.8} & $   62.1$ \small{$\pm$ 1.1} \\

\midrule \\[-2.7ex]
CLIP Frozen                     & $  68.0$ \small{$\pm$ 4.0} & $   9.3$ \small{$\pm$ 2.3} & $   80.0$ \small{$\pm$ 4.0} & $   98.7$ \small{$\pm$ 2.3} & $   56.0$ \small{$\pm$ 8.0} & $   40.0$ \small{$\pm$ 6.9} & $   28.0$ \small{$\pm$ 4.0} & $  43.8$ \small{$\pm$ 4.6} & $   14.9$ \small{$\pm$ 1.0} & $   50.5$ \small{$\pm$ 2.1} & $   63.6$ \small{$\pm$ 2.4} & $   14.4$ \small{$\pm$ 1.1} \\

CLIP Finetuned                   & $  76.0$ \small{$\pm$ 4.0} & $   18.7$ \small{$\pm$ 2.3} & $   90.7$ \small{$\pm$ 2.3} & $   97.3$ \small{$\pm$ 2.3} & $   46.7$ \small{$\pm$ 16.2} & $   65.3$ \small{$\pm$ 12.9} & $   44.0$ \small{$\pm$ 6.9} & $  83.8$ \small{$\pm$ 3.3} & $   49.8$ \small{$\pm$ 9.0} & $   79.7$ \small{$\pm$ 4.3} & $   68.9$ \small{$\pm$ 0.9} & $   31.9$ \small{$\pm$ 5.4} \\

CLIP Finetuned + {\model}       & $  76.0$ \small{$\pm$ 4.0} & $   29.3$ \small{$\pm$ 8.3} & $   98.7$ \small{$\pm$ 2.3} & $   100.0$ \small{$\pm$ 0.0} & $   76.0$ \small{$\pm$ 0.0} & $   90.7$ \small{$\pm$ 8.3} & $   78.7$ \small{$\pm$ 4.6} & $  92.6$ \small{$\pm$ 5.2} & $   77.6$ \small{$\pm$ 3.5} & $   72.0$ \small{$\pm$ 3.0} & $   70.6$ \small{$\pm$ 1.1} & $   42.7$ \small{$\pm$ 5.7} \\

\midrule \\[-2.7ex]
MVP Frozen                     & $  77.3$ \small{$\pm$ 2.3} & $   38.7$ \small{$\pm$ 4.6} & $   92.0$ \small{$\pm$ 6.9} & $   100.0$ \small{$\pm$ 0.0} & $   85.3$ \small{$\pm$ 4.6} & $   80.0$ \small{$\pm$ 4.0} & $   90.7$ \small{$\pm$ 9.2} & $  82.6$ \small{$\pm$ 5.7} & $   52.6$ \small{$\pm$ 7.6} & $   91.7$ \small{$\pm$ 4.9} & $   70.4$ \small{$\pm$ 0.3} & $   25.8$ \small{$\pm$ 7.3} \\

MVP Finetuned                   & $  81.3$ \small{$\pm$ 2.3} & $   82.7$ \small{$\pm$ 8.3} & $  100.0$ \small{$\pm$ 0.0} & $   100.0$ \small{$\pm$ 0.0} & $   89.3$ \small{$\pm$ 4.6} & $   89.3$ \small{$\pm$ 9.2} & $   92.0$ \small{$\pm$ 10.6} & $  96.3$ \small{$\pm$ 0.9} & $   89.8$ \small{$\pm$ 1.3} & $   84.4$ \small{$\pm$ 4.2} & $   69.9$ \small{$\pm$ 0.8} & $   46.7$ \small{$\pm$ 2.1} \\

MVP Finetuned + {\model}        & $  78.7$ \small{$\pm$ 2.3} & $   88.0$ \small{$\pm$ 6.9} & $  100.0$ \small{$\pm$ 0.0} & $   100.0$  \small{$\pm$ 0.0} & $   92.0$ \small{$\pm$ 8.0} & $   86.7$ \small{$\pm$ 2.3} & $   96.0$ \small{$\pm$ 6.9} & $  96.3$ \small{$\pm$ 0.8} & $   88.9$ \small{$\pm$ 2.1} & $   98.0$ \small{$\pm$ 0.5} & $   68.9$ \small{$\pm$ 3.0} & $   50.7$ \small{$\pm$ 6.2} \\

\midrule \\[-2.7ex]
VC-1 Frozen                    & $  73.3$ \small{$\pm$ 4.6} & $   26.7$ \small{$\pm$ 6.1} & $   96.0$ \small{$\pm$ 4.0} & $  100.0$ \small{$\pm$ 0.0} & $   81.3$ \small{$\pm$ 10.1} & $   70.7$ \small{$\pm$ 4.6} & $   85.3$ \small{$\pm$ 8.3} & $  75.5$ \small{$\pm$ 1.7} & $   44.6$ \small{$\pm$ 3.3} & $   83.1$ \small{$\pm$ 5.7} & $   70.4$ \small{$\pm$ 1.3} & $   31.3$ \small{$\pm$ 4.0} \\
VC-1 Finetuned                  & $  74.7$ \small{$\pm$ 2.3} & $   72.0$ \small{$\pm$ 8.0} & $   98.7$ \small{$\pm$ 2.3} & $  100.0$ \small{$\pm$ 0.0} & $   93.3$ \small{$\pm$ 4.6} & $   85.3$ \small{$\pm$ 2.3} & $   92.0$ \small{$\pm$ 10.6} & $  96.5$ \small{$\pm$ 0.5} & $   78.4$ \small{$\pm$ 3.6} & $   83.4$ \small{$\pm$ 8.2} & $   69.6$ \small{$\pm$ 0.9} & $   46.6$ \small{$\pm$ 4.3} \\
VC-1 Finetuned + {\model}       & $  80.0$ \small{$\pm$ 4.0} & $   74.7$ \small{$\pm$ 6.1} & $  100.0$ \small{$\pm$ 0.0} & $  100.0$ \small{$\pm$ 0.0} & $   93.3$ \small{$\pm$ 2.3} & $   90.7$ \small{$\pm$ 2.3} & $   94.7$ \small{$\pm$ 6.1} & $  95.9$ \small{$\pm$ 2.1} & $   86.6$ \small{$\pm$ 3.3} & $   93.6$ \small{$\pm$ 6.8} & $   69.3$ \small{$\pm$ 0.6} & $   58.3$ \small{$\pm$ 8.3} \\

\bottomrule
\end{tabular}}
\end{center}
\label{table: full_main_result}
\end{table}

\section{Adapter-based Methods Results}

\begin{table}[h!]
\caption{\textbf{Detailed full finetuning performance results against adapter methods.} 
We report the full performance results of {\model} compared to adapter-based methods for full finetuning, using CLIP and VC-1 with ViT-B across three independent seeds.}
\begin{center}
\resizebox{0.7\textwidth}{!}{
\begin{tabular}{lcccccc}
\toprule
Model & Module & \# trainable params & Adroit & MetaWorld & DMC &  Mean \\
\midrule \\[-2.5ex]

& X & 85.8M &  47.3 \small{$\pm$ 3.2} & 68.8 \small{$\pm$ 8.1} & 62.8 \small{$\pm$ 4.6} & 59.6 \\ 

\multirow{3}{*}{CLIP} & AdaptFormer$_{64}$ & +1.2M &  48.0 \small{$\pm$ 5.4} & 71.7 \small{$\pm$ 8.0} & 58.6 \small{$\pm$ 3.8} & 59.4 \\ 
& RoboAdapter$_{64}$ & +1.2M & 47.3 \small{$\pm$ 3.2} & 68.8 \small{$\pm$ 5.9} & 61.4 \small{$\pm$ 3.0}
& 59.2 \\
& RoboAdapter$_{192}$ & +3.5M & 47.3 \small{$\pm$ 1.2}  & 67.5 \small{$\pm$ 8.9} & 62.4 \small{$\pm$ 3.7}  & 59.1 \\
& \cellcolor{gray!25}{\model} & \cellcolor{gray!25}+3.1M & \cellcolor{gray!25}\textbf{52.7 \small{$\pm$ 6.2}} & \cellcolor{gray!25}\textbf{88.8 \small{$\pm$ 3.1}} & \cellcolor{gray!25}\textbf{71.1 \small{$\pm$ 3.7}} & \cellcolor{gray!25}\textbf{70.9} \\

\midrule

& X & 85.8M &  73.3 \small{$\pm$ 5.2} & 93.9 \small{$\pm$ 4.0} & 74.9 \small{$\pm$ 3.5} & 80.7 \\ 

\multirow{3}{*}{VC-1} & AdaptFormer$_{64}$ & +1.2M &  75.3 \small{$\pm$ 3.1} & 94.1 \small{$\pm$ 4.3} & 77.4 \small{$\pm$ 2.7} & 82.3 \\ 
& RoboAdapter$_{64}$ & +1.2M & 74.7 \small{$\pm$ 4.3} & 94.9 \small{$\pm$ 2.9} & 79.5 \small{$\pm$ 2.1}
& 83.0 \\
& RoboAdapter$_{192}$ & +3.5M & 75.3 \small{$\pm$ 6.2}  & 95.2 \small{$\pm$ 4.6} & 77.1 \small{$\pm$ 3.2}  & 82.5 \\
& \cellcolor{gray!25}{\model} & \cellcolor{gray!25}+3.1M &  \cellcolor{gray!25}\textbf{77.3} \small{$\pm$ 5.1} & \cellcolor{gray!25}\textbf{95.7} \small{$\pm$ 2.2} & \cellcolor{gray!25}\textbf{80.7} \small{$\pm$ 4.2}  & \cellcolor{gray!25}\textbf{84.6}\\

\bottomrule 
\end{tabular}}
\end{center}
\label{table: fullfinetune_appendix}
\end{table}

\begin{table}[h!]
\caption{\textbf{Detailed PEFT performance results against adapter methods.} 
We report the full performance results of {\model} compared to adapter-based methods for parameter-efficient finetuning, using CLIP and VC-1 with ViT-B across three independent seeds.}
\begin{center}
\resizebox{0.7\textwidth}{!}{
\begin{tabular}{lcccccc}
\toprule
Model & Module & \# trainable params & Adroit & MetaWorld & DMC &  Mean \\
\midrule \\[-2.5ex]

& X & -- &  38.7 \small{$\pm$ 3.2} & 60.5 \small{$\pm$ 5.1} & 37.4 \small{$\pm$ 2.3} & 45.5 \\ 

\multirow{3}{*}{CLIP} & AdaptFormer$_{64}$ & 1.2M &  41.3 \small{$\pm$ 2.3} & 67.7 \small{$\pm$ 8.4} & 45.0 \small{$\pm$ 4.4} &  51.3\\ 
& RoboAdapter$_{64}$ & 1.2M &  44.7 \small{$\pm$ 6.5} & 61.9 \small{$\pm$ 6.5} & 43.7 \small{$\pm$ 3.2} & 50.1 \\
& RoboAdapter$_{192}$ & 3.6M & 45.3 \small{$\pm$ 5.8}  & 61.9 \small{$\pm$ 4.8} & 43.8 \small{$\pm$ 2.6} &  50.3 \\
& \cellcolor{gray!25}{\model} & \cellcolor{gray!25}3.3M & \cellcolor{gray!25}\textbf{51.3\small{$\pm$ 10.1}} & \cellcolor{gray!25}\textbf{86.9 \small{$\pm$ 5.5}} & \cellcolor{gray!25}\textbf{65.4 \small{$\pm$ 4.4}} & \cellcolor{gray!25}\textbf{67.9} \\

\midrule

& X & -- & 50.0 \small{$\pm$ 5.4} & 86.7 \small{$\pm$ 5.4} &  61.0\small{$\pm$ 3.2} & 65.9 \\ 

\multirow{3}{*}{VC-1} & AdaptFormer$_{64}$ & 1.2M &  68.7 \small{$\pm$ 8.4} & 90.9 \small{$\pm$ 5.4} &  76.8 \small{$\pm$ 2.9} & 78.8 \\ 
& RoboAdapter$_{64}$ & 1.2M & 70.0 \small{$\pm$ 4.2} & 93.1 \small{$\pm$ 4.9} & 73.3 \small{$\pm$ 3.0} & 78.8 \\
& RoboAdapter$_{192}$ & 3.6M & 69.3 \small{$\pm$ 5.1}  &  91.5 \small{$\pm$ 5.8} & 75.8 \small{$\pm$ 3.9} & 78.9 \\
& \cellcolor{gray!25}{\model} & \cellcolor{gray!25}3.3M & \cellcolor{gray!25}\textbf{66.7 \small{$\pm$ 8.1}} & \cellcolor{gray!25}\textbf{93.9 \small{$\pm$ 3.1}} & \cellcolor{gray!25}\textbf{77.8 \small{$\pm$ 3.9}} & \cellcolor{gray!25}\textbf{79.5} \\

\bottomrule 
\end{tabular}}
\end{center}
\label{table: peft_appendix}
\end{table}

\newpage

\section{Ablation Results}

\begin{table}[ht]
\caption{\textbf{Detailed performance results for {\model} with larger scale.} We provide the full performance results of {\model} when paired with ViT-B/16 and ViT-L/14 for VC-1 across three independent seeds. We observe that {\model} boosts visuo-motor control task performance for both scales, where ViT-B + {\model} outperforms the larger ViT-L.}
\begin{center}
\resizebox{0.8\textwidth}{!}{
\begin{tabular}{lcccccc}
\toprule
\multirow{2}{*}{Model \& Strategy} & Backbone Scale \& & \multirow{2}{*}{\# trainable params} &  \multirow{2}{*}{Adroit} &  \multirow{2}{*}{MetaWorld} & \multirow{2}{*}{DMC} & \multirow{2}{*}{Mean} \\
& Additional Module &  &  \\
\midrule \\[-2.5ex]

\multirow{3}{*}{VC-1+} & ViT-B & 85.8M &  73.3 \small{$\pm 5.2$} & 93.9 \small{$\pm 4.0$} & 74.9 \small{$\pm 3.5$} &  80.7\\ 
 \multirow{3}{*}{Finetuned} & \cellcolor{gray!25}ViT-B + {\model} & \cellcolor{gray!25}88.9M &   \cellcolor{gray!25}77.3 \small{$\pm 5.1$} &   \cellcolor{gray!25}95.7 \small{$\pm 2.2$} &   \cellcolor{gray!25}80.7\small{$\pm 4.2$} &   \cellcolor{gray!25}84.6 \\
\cline{2-7}
\\[-1.7ex]
& ViT-L  & 303.3M &  78.7 \small{$\pm 7.6$} &  95.2\small{$\pm 4.9$} &  76.3\small{$\pm 1.1$} & 83.4 \\
 & \cellcolor{gray!25}ViT-L + {\model} & \cellcolor{gray!25}307.7M & \cellcolor{gray!25}76.0 \small{$\pm 6.1$} & \cellcolor{gray!25}97.6 \small{$\pm 3.1$} & \cellcolor{gray!25}79.8 \small{$\pm 3.3$} & \cellcolor{gray!25}84.5 \\

\bottomrule 
\end{tabular}}
\end{center}
\label{table: scaling_full_table}
\vskip -0.1in
\end{table}

\section{Additional Visualization of {\model}}
\label{appendix: visualization}

\begin{figure*}[t]
\begin{center}
\includegraphics[width=0.75\textwidth]{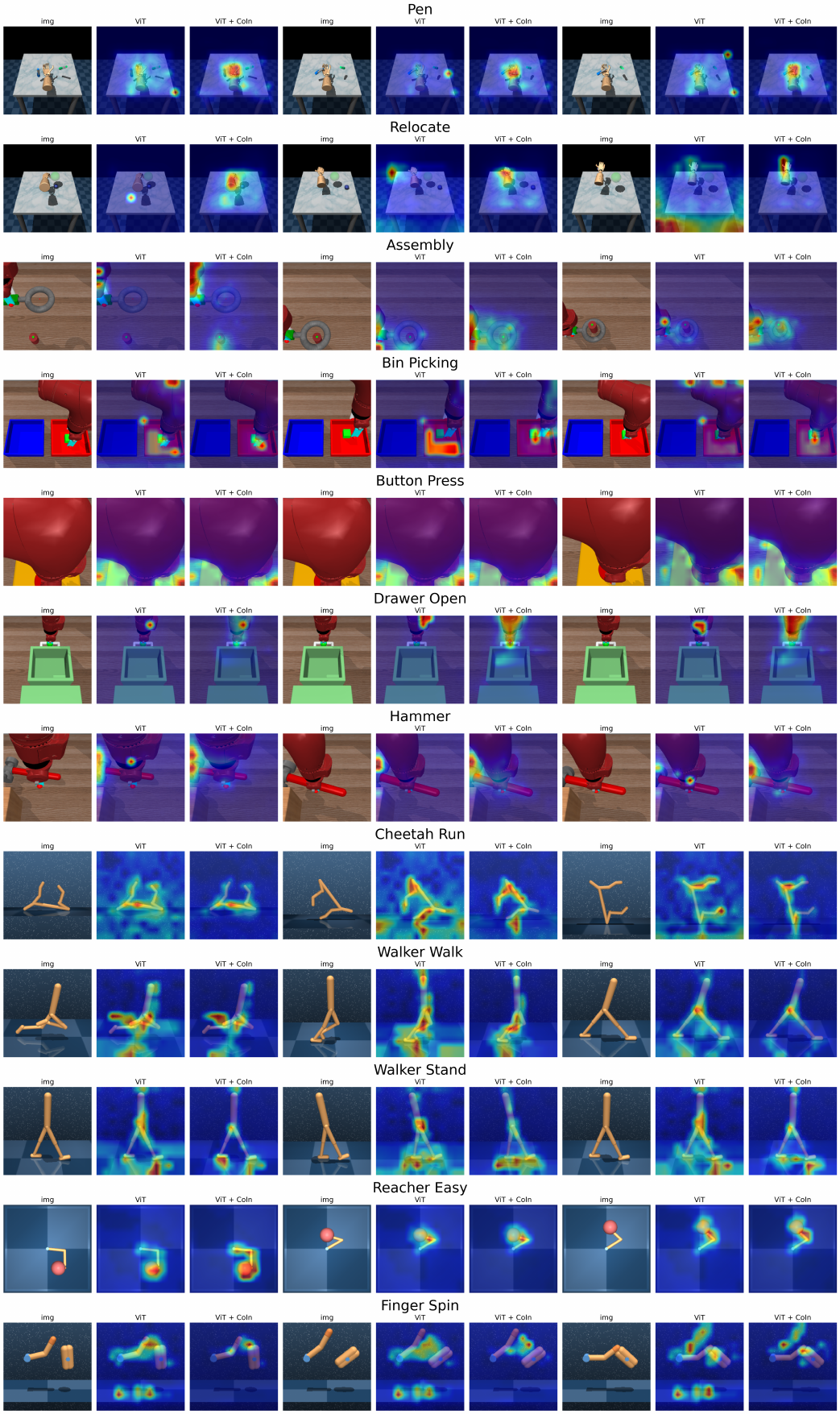}
\end{center}
\vspace{-2mm}
\caption{\textbf{Attention rollout visualization.} Additional qualitative attention map visualization for all tasks.}
\label{fig:additional_qualitative}
\end{figure*}

To analyze where our model focuses on the input images, we apply the Attention Rollout technique as described by \citet{abnar2020attentionrollout, vitattentionrollout}. We first select the attention heads with the maximum attention weights (minimum for DMC) and eliminate 90\% of the attention pixels to concentrate on the most significant parts. Figure~\ref{fig:additional_qualitative} presents further qualitative examples of these attention rollouts. Overall, the integration of ViT with {\model} demonstrates a relatively precise ability to identify the positions of hands/grippers and objects (Adroit and MetaWorld) as well as the agents (DMC).

\end{document}